\documentclass{tlp}
\usepackage{aopmath}

\newtheorem{definition}{Definition} 
\newtheorem{example}{Example} 

\begin{document}
\bibliographystyle{tlp}

\long\def\comment#1{}

\title{Two Results for Prioritized Logic Programming}

\author[Y. Zhang]
{YAN ZHANG \\
School of Computing and Information Technology\\
University of Western Sydney\\
Locked Bag 1797, Penrith South DC\\ 
NSW 1797, Australia    \\
E-mail: yan@cit.uws.edu.au
}

\pagerange{\pageref{firstpage}--\pageref{lastpage}}
\volume{\textbf{10} (3):}
\jdate{March 2002}
\setcounter{page}{1}
\pubyear{2002}

\maketitle

\label{firstpage}

\begin{abstract}
Prioritized default reasoning has illustrated its rich expressiveness and
flexibility in knowledge representation and reasoning.
However, many important aspects of 
prioritized default reasoning have yet to be thoroughly explored.
In this paper, we  
investigate two properties of 
prioritized logic programs in the context of answer set semantics.
Specifically, we reveal a close relationship between mutual defeasibility
and uniqueness of the answer set for a prioritized logic program.
We then explore how the splitting technique for 
extended logic programs can be extended to
prioritized logic programs. We prove splitting theorems
that can be used to simplify the evaluation of a prioritized logic program 
under certain conditions.
%
%
\end{abstract}

\section{Introduction}

Prioritized default reasoning has illustrated its rich expressiveness and
flexibility in knowledge representation, reasoning about action and
rule updates \cite{brewka:plp,g:97,yan:plp}.
Recently, different approaches and
formulations for prioritized default reasoning based on
logic programming and default theories have been proposed
\cite{be:aij,d:plp,g:97,w:plp}. 
However, most of these proposals mainly focus on the semantic development,
while other important
properties are usually not thoroughly explored.

In this paper, we
investigate two specific properties of
prioritized logic programs in the context of answer set semantics.
First, we reveal a close relationship between mutual defeasibility
and uniqueness of the answer set for a prioritized logic program.
Mutual defeasibility can be viewed as a way of characterizing rules in
a logic program, where two rules in the program are mutually defeasible if
triggering one rule may cause a defeat of the other, and {\em vice versa}.
It is quite easy to observe that a logic program does not contain a pair of
mutually defeasible rules if this program is locally stratified. However, the converse
does not hold. We then 
provide a sufficient condition to ensure the uniqueness of the answer set for
a prioritized logic program.
We show that our characteristic condition is weaker than
the traditional local stratification for general logic programs \cite{apt:94}.

Second, we investigate the splitting technique for 
prioritized logic programs. It is well known that
Lifschitz and Turner's Splitting Set Theorem \cite{lt:lp94}
for extended logic programs
may significantly simplify the computation of
answer sets of an extended logic program.
The basic idea of splitting technique 
is that under certain conditions, an extended logic
program can be split into several ``smaller components'' such that
the computation of the answer set of the original 
program is reduced to the computation of the answer set of these smaller 
components. We show that this splitting technique 
is also suitable for computing answer sets of prioritized logic programs.
Furthermore, our splitting theorems for prioritized logic programs
also provide a generalization of Lifschitz and Turner's result. 

The paper is organized as follows. Section 2 develops  
the syntax and semantics of prioritized logic programs. In our formulation,
a prioritized logic program is defined to be an extended logic program associating 
with a strict partial ordering on rules in the program. An answer set semantics 
for prioritized logic programs
is then defined. Several basic properties of prioritized logic programs are
also studied in this section. By introducing the concept of mutual 
defeasibility, section 3 proves a sufficient condition to characterize
the unique answer set for a prioritized logic program.
Section 4 then extends the splitting technique for extended logic programs to
prioritized logic programs.
Finally, section 5 concludes the paper with some remarks.

\section{Prioritized Logic Programs}

To specify prioritized logic programs (PLPs), we first
introduce the extended logic program and its answer set semantics
developed by Gelfond and Lifschitz \cite{gl:logic}.
A language $\cal L$ of extended
logic programs is determined by its object constants, function
constants and predicate constants. {\em Terms} are
built as in the corresponding first order language; {\em atoms}
have the form $P(t_{1},\cdots,t_{n})$, where
$t_{i}$ ($1\leq i\leq n$) is a term and $P$ is a predicate constant of arity $n$;
a {\em literal} is either an atom $P(t_{1},\cdots,t_{n})$ or
a negative atom $\neg P(t_{1},\cdots,t_{n})$.
A {\em rule} is an expression of the form:
\begin{equation}
L_{0}\leftarrow L_{1},\cdots,L_{m}, not L_{m+1},\cdots, not L_{n},
\label{eq2}
\end{equation}
where each $L_{i}$ ($0\leq i\leq n$) is a literal.
$L_{0}$ is called the {\em head} of the rule,
while $\{L_{1},\cdots,L_{m}$, $not L_{m+1},\cdots$,
$not L_{n}\}$ is called the {\em body} of the rule.
Obviously, the body of a
rule could be empty. We also allow the head of a rule to be empty. In 
this case, the rule with an empty head is called {\em constraint}. 
A term, atom, literal, or rule is {\em ground} if no variable
occurs in it.
An {\em extended logic program} $\Pi$ is a collection of rules.
$\Pi$ is {\em ground} if each rule in $\Pi$ is
ground. 

Let $r$ be a ground rule of the form (\ref{eq2}), 
we use $pos(r)$ to denote the set of literals in the body of $r$
without negation as failure
$\{L_{1},\cdots,L_{m}\}$, and $neg(r)$ the set of literals
in the body of $r$ with negation as failure $\{L_{m+1},\cdots,L_{n}\}$. We specify
$body(r)$ to be
$pos(r)\cup neg(r)$. We also use $head(r)$ to denote the head of $r$: $\{L_{0}\}$.
Then we use $lit(r)$ to denote $head(r)\cup body(r)$. By extending these
notations, we use $pos(\Pi)$, $neg(\Pi)$,
$body(\Pi)$, $head(\Pi)$, and $lit(\Pi)$ to denote the unions of
corresponding components of all rules in the ground program $\Pi$, e.g.
$body(\Pi)=\bigcup_{r\in\Pi} body(r)$. If $\Pi$ is a non-ground program,
then notions $pos(\Pi)$, $neg(\Pi)$,
$body(\Pi)$, $head(\Pi)$, and $lit(\Pi)$ are defined based on the ground
instantiation (see below definition) of $\Pi$.

To evaluate an extended logic program, Gelfond and
Lifschitz proposed an answer set semantics for extended logic
programs. 
Let $\Pi$ be a ground extended logic program not containing {\em not}
and {\em Lit} the set of all ground literals in the language of $\Pi$.
An {\em answer set} of $\Pi$ 
is the smallest subset $S$ of {\em Lit} such that
(i) for any rule $L_{0}\leftarrow L_{1},\cdots,L_{m}$ from
$\Pi$, if $L_{1},\cdots,L_{m}\in S$, then $L_{0}\in S$; and (ii)
if $S$ contains a pair of complementary literals, then
$S=Lit$.
Now let $\Pi$ be a ground arbitrary extended logic program. For any subset $S$
of $Lit$, let $\Pi^{S}$ be the logic program obtained
from $\Pi$ by deleting
(i) each rule that has a formula
{\em not} $L$ in its body with $L\in S$, and
(ii) all formulas of the form {\em not} $L$ in the bodies
of the remaining rules\footnote{We also
call $\Pi^{S}$ the Gelfond-Lifschitz transformation of
$\Pi$ in terms of $S$.}.
We define that $S$ is an {\em answer set} of $\Pi$
iff $S$ is an answer set of $\Pi^{S}$.

For a non-ground extended logic program $\Pi$, we usually view a rule
in $\Pi$ containing variables to be the set of all ground instances of this
rule formed from the set of ground literals in the language. The 
collection of all these ground rules forms the 
{\em ground instantiation} $\Pi'$ of $\Pi$. 
Then a set of ground literals is an answer set of
$\Pi$ if and only if it is an answer set of $\Pi'$.
It is easy to see that an
extended logic program may have one, more than one, or
no answer set at all.

A {\em prioritized logic program} (PLP) ${\cal P}$ is a triple
$(\Pi,{\cal N},<)$, where
$\Pi$ is an extended logic program, $\cal N$ is a
naming function
mapping each rule in $\Pi$ to a name, and $<$ is a strict
partial ordering on names.
The partial ordering $<$ in ${\cal P}$ plays an essential role in
the evaluation of ${\cal P}$.
We also use ${\cal P}(<)$ to denote the set of $<$-relations of ${\cal P}$. 
Intuitively $<$ represents a preference
of applying rules during the evaluation of the program.
In particular, if
${\cal N}(r)<{\cal N}(r')$ holds in ${\cal P}$, rule $r$ would be
preferred to apply over rule $r'$ during the evaluation of ${\cal P}$
(i.e. rule $r$ is more preferred than rule $r'$).
Consider the following
classical example represented in our formalism:
\begin{quote}
${\cal P}_{1}=(\Pi,{\cal N},<)$:\\
\hspace*{.1in} $N_{1}: Fly(x)\leftarrow Bird(x)$, {\em not} $\neg Fly(x)$,\\
\hspace*{.1in} $N_{2}: \neg Fly(x)\leftarrow Penguin(x)$, {\em not} $Fly(x)$, \\
\hspace*{.1in} $N_{3}: Bird(Tweety)\leftarrow$,\\
\hspace*{.1in} $N_{4}: Penguin(Tweety)\leftarrow$,\\
\hspace*{.1in} $N_{2}<N_{1}$.
\end{quote}
Obviously, rules $N_{1}$ and $N_{2}$ conflict with each other
as their heads are complementary literals,
and applying $N_{1}$ will defeat
$N_{2}$ and {\em vice versa}.
However, as $N_{2}<N_{1}$, we would expect that rule $N_{2}$ is preferred
to apply first and then defeat rule $N_{1}$ so that
the desired solution $\neg Fly(Tweety)$ can be derived.

\begin{definition}
Let $\Pi$ be a ground extended logic program and $r$ a ground rule of the
form (\ref{eq2}) 
($r$ does not necessarily belong to $\Pi$).
Rule $r$ is {\em defeated}
by $\Pi$ iff $\Pi$ has an answer set and
for any answer set $S$ of $\Pi$,
there exists some $L_{i}\in S$, where $m+1\leq i\leq n$.
\end{definition}

Now our idea of evaluating a PLP is as follows.
Let ${\cal P}=(\Pi,{\cal N},<)$. If there are two rules
$r$ and $r'$ in $\Pi$ and ${\cal N}(r)<{\cal N}(r')$,
$r'$ will be ignored in the evaluation of ${\cal P}$, {\em only if}
keeping $r$ in $\Pi$ and deleting $r'$
from $\Pi$ will result in a defeat of $r'$.
By eliminating all such potential rules from $\Pi$,
$\cal P$ is eventually reduced to an extended logic program in which
the partial ordering $<$ has been removed. Our evaluation for
${\cal P}$ is then based on this {\em reduced} extended logic program.

Similarly to the case of extended logic programs, 
the evaluation of a PLP will be based on its ground form. 
We say that a PLP ${\cal P'}=(\Pi',{\cal N'},<')$ is the
{\em ground instantiation} of ${\cal P}=(\Pi,{\cal N},<)$ if 
(1) $\Pi'$ is the ground instantiation of $\Pi$; and (2) 
${\cal N'}(r_1')<' {\cal N'}(r_2')\in {\cal P'}(<')$ if and only
if there exist rules $r_1$ and $r_2$ in $\Pi$ such that
$r_1'$ and $r_2'$ are ground instances of $r_1$ and $r_2$ respectively and
${\cal N}(r_1) < {\cal N}(r_2)\in {\cal P}(<)$.
Under this definition, 
however, we require a restriction on a PLP since
not every PLP's ground instantiation presents a consistent
information with respect to the original PLP.
Consider a PLP as follows:
\begin{quote}
$N_{1}: P(f(x))\leftarrow not P(x)$,\\
\hspace*{.1in} $N_{2}: P(f(f(x)))\leftarrow not P(f(x))$,\\
\hspace*{.1in} $N_{2}<N_{1}$.
\end{quote}
If the only constant in the language is $0$, then the set of ground
instances of $N_{1}$ and $N_{2}$ includes rules like:
\begin{quote}
$N_{1}': P(f(0))\leftarrow not P(0)$,\\
\hspace*{.1in} $N_{2}': P(f(f(0)))\leftarrow not P(f(0))$,\\
\hspace*{.1in} $N_{3}': P(f(f(f(0))))\leftarrow not P(f(f(0)))$,\\
\hspace*{.1in} $\cdots$,
\end{quote}
It is easy to see that $N_{2}'$ can be viewed as an
instance for both $N_{1}$ and $N_{2}$. Therefore, the ordering $<'$
among rules $N_1', N_2',N_3',\cdots$
is no longer a partial ordering because of
$N_{2}'<'N_{2}'$.
Obviously, we need to exclude this kind of programs in our context.
On the other hand,
we also want to avoid a situation like
$\cdots <'N_{3}'<'N_{2}'<'N_{1}'$ in the ground
prioritized logic program because this $<'$ indicates that
there is no most preferred rule in the program.

Given a PLP ${\cal P}=(\Pi, {\cal N}, <)$. We say that
${\cal P}$ is {\em well formed} if
there is no rule $r'$ that is an instance of two different rules
$r_{1}$ and $r_{2}$ in $\Pi$ and
${\cal N}(r_{1})<{\cal N}(r_{2}) \in {\cal P}(<)$.
Then it is not difficult to observe that
the following fact holds.

\begin{quote}
{\bf Fact}: If a PLP ${\cal P}=(\Pi, {\cal N}, <)$ is well formed, then
in its ground instantiation ${\cal P'}=(\Pi', {\cal N'}, <')$, $<'$ is
a partial ordering and
every non-empty subset of $\Pi'$ has a least element with respect to $<'$.
\end{quote}

Due to the above fact,
in the rest of this paper, we will only consider well formed PLP programs
in our
discussions, and consequently, the evaluation for an arbitrary
PLP ${\cal P}=(\Pi,{\cal N},<)$  will be based on its ground
instantiation ${\cal P'}=(\Pi', {\cal N'}, <')$. Therefore, in our context
a ground prioritized (or extended) logic program may contain
infinite number of rules. In this case, we will assume that
this ground program is the ground instantiation of
some program that only contains
finite number of rules. In the rest of the paper, whenever
there is no confusion, we will only consider
ground prioritized (extended) logic programs without explicit 
declaration.

\begin{definition}
\cite{yan:plp}
Let ${\cal P}=(\Pi,{\cal N},<)$ be a prioritized logic program.
${\cal P}^{<}$ is a {\em reduct} of $\cal P$ with respect to $<$
if and only if there exists a sequence of sets $\Pi_{i}$
($i=0, 1, \cdots$) such that:
\begin{enumerate}
\item $\Pi_{0}=\Pi$;
\item $\Pi_{i}=\Pi_{i-1}-\{r_{1},r_{2},\cdots \mid$ (a)
there exists $r\in \Pi_{i-1}$ such that \\
\hspace*{.2in} for every $j$ ($j=1,2, \cdots$),
${\cal N}(r)<{\cal N}(r_{j})\in {\cal P}(<)$ and \\
\hspace*{.2in} $r_{1}, r_2, \cdots$
are defeated by $\Pi_{i-1}-\{r_{1}, r_{2}, \cdots\}$, and (b) there \\
\hspace*{.2in} are no rules $r',r'', \cdots \in\Pi_{i-1}$
such that $N(r_{j})<N(r')$, \\
\hspace*{.2in} $N(r_{j})<N(r''),\cdots$ 
for some $j$ ($j=1, 2, \cdots$)
and $r', r'', \cdots$ \\
\hspace*{.2in} are defeated by $\Pi_{i-1}-\{r', r'', \cdots\} \}$;
\item ${\cal P}^{<}=\bigcap_{i=0}^{\infty} \Pi_{i}$.
\end{enumerate}
\end{definition}

In Definition 2,
${\cal P}^{<}$ is an extended logic program
obtained from
$\Pi$ by eliminating some rules from $\Pi$.  In particular,
if ${\cal N}(r)<{\cal N}(r_{1})$, ${\cal N}(r)<{\cal N}(r_{2})$,
$\cdots$,
and $\Pi_{i-1}-\{r_{1}, r_{2},\cdots\}$ defeats $\{r_{1}, r_{2},\cdots\}$, then rules
$r_{1},r_{2},\cdots$
will be eliminated from $\Pi_{i-1}$ if no {\em less preferred rule} can be
eliminated (i.e. conditions (a) and (b)).
This procedure is continued until a fixed point is reached.
It should be noted that condition (b) in the above definition is 
necessary because without it some unintuitive 
results may be derived. For instance, consider ${\cal P}_{1}$ again,
if we add additional preference $N_3<N_2$ in ${\cal P}_{1}$, then using
a modified version of Definition 2 without condition (b), 
\begin{quote}
$\{Fly(Tweety)\leftarrow Bird(Tweety), not \neg Fly(Tweety)$, \\
\hspace*{.2in} $Bird(Tweety)\leftarrow$,\\
\hspace*{.2in} $Penguin(Tweety)\leftarrow\}$
\end{quote} 
is a reduct of ${\cal P}_1$, from which we will 
conclude that Tweety can fly.

\begin{definition}
\cite{yan:plp}
Let ${\cal P}=(\Pi,{\cal N},<)$ be a PLP and $Lit$ the set of all
ground literals in the language of $\cal P$. For any subset $S$ of $Lit$, $S$
is an {\em answer set} of $\cal P$ iff
$S$ is an answer set for some reduct ${\cal P}^{<}$ of ${\cal P}$.
\end{definition}

Using Definitions 2 and 3, it is easy to conclude that ${\cal P}_{1}$
has a unique reduct as follows:
\begin{quote}
${\cal P}_{1}^{<}=\{\neg Fly(Tweety)\leftarrow Penguin(Tweety)$, 
{\em not} $Fly(Tweety)$,\\
\hspace*{.5in} $Bird(Tweety)\leftarrow$, \\
\hspace*{.5in} $Penguin(Tweety)\leftarrow\}$,
\end{quote}
from which we obtain the following answer set of ${\cal P}_{1}$:
\begin{quote}
$S=\{Bird(Tweety)$, $Penguin(Tweety)$, $\neg Fly(Tweety)\}$.
\end{quote}

Now we consider another program ${\cal P}_{2}$:

\begin{quote}
$N_{1}: A\leftarrow$,\\
\hspace*{.1in} $N_{2}: B\leftarrow$ {\em not} $C$,\\
\hspace*{.1in} $N_{3}: D\leftarrow$,\\
\hspace*{.1in} $N_{4}: C\leftarrow$ {\em not} $B$,\\
\hspace*{.1in} $N_{1}<N_{2}, N_{3}<N_{4}$.
\end{quote}

According to Definition 2, it is easy to see that
${\cal P}_{2}$ has two reducts:
\begin{quote}
$\{A\leftarrow$, \hspace*{.05in}
$D\leftarrow$, 
\hspace*{.05in} $C\leftarrow$ {\em not} $B\}$, and\\
\hspace*{.1in}  $\{A\leftarrow$, 
\hspace*{.05in} $B\leftarrow$ {\em not} $C$,\
\hspace*{.05in} $D\leftarrow\}$.
\end{quote}
From Definition 3, it follows that ${\cal P}_{2}$ has two answer sets:
$\{A,C,D\}$ and $\{A,B,D\}$.

To see whether our PLP semantics gives intuitive results in
prioritized default reasoning, we further consider a program
${\cal P'}_2$ - a variation of program ${\cal P}_2$, as follows.
\begin{quote}
$N_1: A\leftarrow$,\\
\hspace*{.1in} $N_2: B\leftarrow$ {\em not} $C$, \\
\hspace*{.1in} $N_3: C\leftarrow$ {\em not} $B$,\\
\hspace*{.1in} $N_1<N_2$.
\end{quote}
It is easy to see that ${\cal P'}_2$ has one answer set $\{A, C\}$.
People may think that this result is not quite intuitive because 
rule $N_2$ is defeated in the evaluation although there is no
preference between $N_2$ and $N_3$. To explain why 
$\{A, C\}$ is a {\em reasonable} answer set of ${\cal P'}_2$, we should
review the concept of defeatness in our formulation (Definition 1).
In a PLP, when we specify one rule is less preferred than the other, for instance,
$N_1<N_2$ ($N_2$ is less preferred than $N_1$), it does not mean
that $N_2$ should be defeated by $N_1$ iff conflict occurs between them. 
Instead, it just means that $N_2$ has a lower priority than $N_1$
to be taken into account in the evaluation of
the {\em whole} program while other rules should be retained in 
the evaluation process if no preference is specified between 
$N_2$ and them. This intuition is captured by
the notion of defeatness in Definition 1 and Definition 2. 

Back to the above example, although there is no direct conflict between
$N_1$ and $N_2$ {\em and} no preference is specified between
$N_2$ and $N_3$ (where conflict exists between them), $N_2$ 
indeed has a lower priority than $N_1$ to be 
applied in the evaluation of ${\cal P'}_2$, which causes
$N_2$ to be defeated.

Now we illustrate several basic properties of
prioritized logic programs. As we mentioned earlier,
when we evaluate a
PLP, a rule including variables is viewed
as the set of its all ground instances.  
Therefore, we are actually dealing with {\em ground} 
prioritized logic programs that may consist of infinite collection 
of rules.
We first introduce some useful notations.
Let $\Pi$ be an extended logic program. We use ${\cal A}(\Pi)$ to
denote the class of all answer sets of $\Pi$.
Suppose ${\cal P}=(\Pi,{\cal N},<)$ is a PLP. From Definition 2, we can see
that a reduct ${\cal P}^{<}$ of ${\cal P}$ is generated
from a sequence of extended logic programs:
$\Pi=\Pi_{0}, \Pi_{1}, \Pi_{2}$, $\cdots$. We use 
$\{\Pi_{i}\}$ ($i=0, 1, 2, \cdots$) to denote this
sequence and call it
a {\em reduct chain} of ${\cal P}$.

\begin{proposition}
Let ${\cal P}=(\Pi,{\cal N},<)$ be a PLP and
$\{\Pi_{i}\}$ ($i=0, 1, 2, \cdots$) its reduct chain. Suppose
$\Pi$ has an answer set.  Then for any $i$ and $j$ where $i<j$,
${\cal A}(\Pi_{j})\subseteq {\cal A}(\Pi_{i})$.
\end{proposition}

\noindent
\begin{proof}
Let $\{\Pi_{i}\}$ ($i=0, 1, 2, \cdots$) be a reduct
chain of ${\cal P}$.
Suppose $S_{j}$ is an answer set of $\Pi_{j}$ for some $j>0$. To prove the
result, it is sufficient to show that
$S_{j}$ is also an answer set of $\Pi_{j-1}$.
According to Definition 2, $\Pi_{j}$ is obtained by eliminating
some rules from $\Pi_{j-1}$ where all these eliminated rules are
defeated by $\Pi_{j}$. So we can express:
\begin{quote}
$\Pi_{j}=\Pi_{j-1}-\{r_{1},r_{2},\cdots\}$.
\end{quote} 
Since $r_{1},r_{2}, \cdots$ are defeated by $\Pi_{j}$, we can write rules
$r_{1},r_{2}, \cdots$ to the following forms:
\begin{quote}
$r_{1}: L_{1}\leftarrow \cdots$, {\em not} $L_{1}', \cdots$,\\
\hspace*{.1in} $r_{2}: L_{2}\leftarrow \cdots$, {\em not} $L_{2}', \cdots$,\\
\hspace*{.1in} $\cdots$,
\end{quote} 
where $L_{1}',L_{2}',\cdots\in S_{j}$. Now consider Gelfond-Lifschitz transformation of
$\Pi_{j}$ in terms of $S_{j}$. It is clear that during the transformation, each
rule in $\Pi_{j}$ including $not$ $L_{1}'$, $not$ $L_{2}'$, $\cdots$ in
its body will be deleted. From here it follows that adding any rule with
$not$ $L_{1}'$ $not$ $L_{2}'$, $\cdots$  in its body will not play any role in the
evaluation of the answer set of the program. So we add
rules $r_{1}, r_{2}, \cdots$ into $\Pi_{j}$, This makes $\Pi_{j-1}$.
Then we have $\Pi_{j}^{S_{j}}=\Pi_{j-1}^{S_{j}}$.
So $S_{j}$ is also an answer set of $\Pi_{j-1}$.
\end{proof}

Proposition 1 shows an important property of the reduct chain
of ${\cal P}$: each $\Pi_{i}$ is consistent
with $\Pi_{i-1}$ but becomes more {\em specific} than
$\Pi_{i-1}$ in the sense
that all answer sets of $\Pi_{i}$ are answer sets of
$\Pi_{i-1}$ but some answer sets of $\Pi_{i-1}$ are
filtered out if they conflict with the preference partial ordering $<$.

\begin{example}
Consider a PLP ${\cal P}_{3}=(\Pi,{\cal N},<)$:
\begin{quote}
$N_{1}: A\leftarrow$ {\em not} $B$,\\
\hspace*{.1in} $N_{2}: B\leftarrow$ {\em not} $A$,\\
\hspace*{.1in} $N_{3}: C\leftarrow$ {\em not} $B$, {\em not} $D$,\\
\hspace*{.1in} $N_{4}: D\leftarrow$ {\em not} $C$,\\
\hspace*{.1in} $N_{1}<N_{2}, N_{3}<N_{4}$.
\end{quote}
From Definition 2, we can see that ${\cal P}_{3}$ has a reduct chain
$\{\Pi_{i}\}$ ($i=0,1,2$):
\begin{quote}
$\Pi_{0}$:\\
\hspace*{.2in} $A\leftarrow$ {\em not} $B$,\\
\hspace*{.2in} $B\leftarrow$ {\em not} $A$,\\
\hspace*{.2in} $C\leftarrow$ {\em not} $B$, {\em not} $D$,\\
\hspace*{.2in} $D\leftarrow$ {\em not} $C$,\\
\hspace*{.1in} $\Pi_{1}$:\\
\hspace*{.2in} $A\leftarrow$ {\em not} $B$,\\
\hspace*{.2in} $C\leftarrow$ {\em not} $B$, {\em not} $D$,\\
\hspace*{.2in} $D\leftarrow$ {\em not} $C$,\\
\hspace*{.1in} $\Pi_{2}$:\\
\hspace*{.2in} $A\leftarrow$ {\em not} $B$,\\
\hspace*{.2in} $C\leftarrow$ {\em not} $B$, {\em not} $D$.
\end{quote}
It is easy to verify that $\Pi_{0}$ has three answer sets
$\{A,C\}$, $\{B,D\}$ and $\{A,D\}$,
$\Pi_{1}$ has two answer sets $\{A,C\}$ and $\{A,D\}$, and
$\Pi_{2}$ has a unique answer set which is also the answer set
of ${\cal P}_{3}$: $\{A,C\}$.
\end{example}

The following theorem shows the answer set relationship between
a PLP and its corresponding extended logic programs.

\begin{theorem}
Let ${\cal P}=(\Pi,{\cal N},<)$ be a PLP and $S$ a subset of $Lit$.
Then the following are equivalent:
\begin{enumerate}
\item $S$ is an answer set of ${\cal P}$.
\item $S$ is an answer set
of each $\Pi_{i}$ for some reduct chain $\{\Pi_{i}\}$ ($i=0,1,2,\cdots$)
of ${\cal P}$.
\end{enumerate}
\end{theorem}

\noindent   
\begin{proof}
(1 $\Rightarrow$ 2) Let ${\cal P}^{<}$ be a reduct of ${\cal P}$ obtained
from a reduct chain $\{\Pi_{i}\}$ ($i=0,1,2,\cdots$)
of ${\cal P}$. By applying Theorem 3 in section 3, it is easy to show that
any reduct chain of ${\cal P}$ is finite.
Therefore,
there exists some $k$ such that $\{\Pi_{i}\}$ ($i=0,1,2,\cdots, k$)
is the reduct chain. This follows that ${\cal P}^{<}=\Pi_k\subseteq \Pi_i$
($i=1,\cdots,k$). 
So from Proposition
1, an answer set of ${\cal P}^{<}$ is also an answer set of $\Pi_{i}$
($i=1,\cdots,k$).
            
(2 $\Rightarrow$ 1) Given a reduct chain
$\{\Pi_{i}\}$ ($i=0,1,2,\cdots$) of
${\cal P}$. 
From the above, since $\{\Pi_{i}\}$ ($i=0,1,2,\cdots$) is finite, 
we can assume that $\{\Pi_{i}\}$ ($i=0,1,2,\cdots, k$)
is the reduct chain. As $\Pi_j\subseteq \Pi_i$ if $j>i$, it 
follows that $\bigcap _{i=0}^{k}\Pi_{i}=\Pi_k$. So the
fact that $S$ is an answer set of
$\Pi_k$ implies that $S$ is also an answer set of ${\cal P}$.
\end{proof} 

\begin{corollary}
If a PLP ${\cal P}=(\Pi,{\cal N},<)$ has an answer set $S$, then $S$ is also
an answer set of $\Pi$.
\end{corollary}

\noindent   
\begin{proof}
From Theorem 1, it shows that if ${\cal P}$ has an answer set
$S$, then $S$ is also an answer set of each $\Pi_{i}$ for
${\cal P}$'s a reduct chain $\{\Pi_{i}\}$ ($i=0, 1, 2, \cdots$),
where $\Pi_{0}=\Pi$. So $S$ is also an answer set of $\Pi$.
\end{proof} 

The following theorem presents a sufficient and necessary
condition for the answer set existence of a PLP.

\begin{theorem}
Let ${\cal P}=(\Pi,{\cal N},<)$ be a PLP.
${\cal P}$ has an answer set if and only if
$\Pi$ has an answer set.
\end{theorem}

\noindent   
\begin{proof}
According to Corollary 1, we only need to prove that if $\Pi$ has an
answer set, then ${\cal P}$ also has an answer set.
Suppose $\Pi$ has an answer set and $\{\Pi_i\}$ ($i=0,1,\cdots$) 
is a reduct chain of ${\cal P}$. From the construction of
$\{\Pi_i\}$ (Definition 2), it is easy to see that every 
$\Pi_i$ ($i=0,1,\cdots$) must have an answer set. On the other hand,
as we have mentioned
in the proof of Theorem 1, 
${\cal P}$'s reduct chain is actually finite: $\{\Pi_i\}$ ($i=0,1,\cdots, k$).
That follows ${\cal P}^{<}=\Pi_k$. Since $\Pi_k$ has an answer set, 
it concludes ${\cal P}$ has an answer set as well.
\end{proof} 
            
\begin{proposition}
Suppose a PLP ${\cal P}$ has a unique reduct. If
${\cal P}$ has a consistent answer set, then ${\cal P}$'s every answer
set is also consistent.
\end{proposition}

\noindent   
\begin{proof}
The fact that ${\cal P}$ has a consistent answer set implies that
${\cal P}$'s reduct ${\cal P}^{<}$ (note
${\cal P}^{<}$ is an extended logic program)
has a consistent answer set. Then from
the result showed in section 2 of \cite{lt:lp94}
(i.e. if an extended logic program has a consistent answer set, then its
every answer set is also consistent), it follows that
${\cal P}^{<}$'s every answer set is also consistent.
\end{proof} 
            
\section{Mutual Defeasibility and Unique Answer Set}

In this section, we try to provide a sufficient condition to 
characterize the uniqueness
of the answer set for a prioritized logic program.
The following definition extends the concept of local stratification for
general logic programs \cite{apt:94,ch:stra,lp:92} to extended logic programs.

\begin{definition}
Let $\Pi$ be an extended logic program 
and $Lit$ be the set of all
ground literals of $\Pi$.
\begin{enumerate}
\item A {\em local stratification} for $\Pi$ is a function
{\em stratum} from $Lit$ to the countable ordinals.
\item Given a local stratification $stratum$, we extend it to ground literals with
negation as failure by setting
$stratum(\mbox{not }L)=stratum(L)+1$, where $L$ is a ground literal.
\item A rule
$L_{0}\leftarrow L_{1},\cdots, L_{m}$, not $L_{m+1},\cdots$, not $L_{n}$
in $\Pi$ is {\em locally stratified} with respect to
$stratum$ if
\begin{quote}
$stratum(L_{0})\geq stratum(L_{i})$, where $1\leq i\leq m$, and\\
$stratum(L_{0})> stratum(not L_{j})$, where $m+1\leq j\leq n$.
\end{quote}
\item $\Pi$ is called {\em locally stratified} with respect to $stratum$
if all of its rules are locally stratified. $\Pi$ is called
{\em locally stratified} if it is locally stratified with respect to
some local stratification.
\end{enumerate}
\end{definition}

It is easy to see that the corresponding extended logic 
program of ${\cal P}_{1}$ (see section 2) is not locally stratified. 
In general, we have the following sufficient condition to
ensure the uniqueness of the answer set for an extended logic program.

\begin{proposition}
Let $\Pi$ be an extended logic program. If $\Pi$ is
locally stratified,
then $\Pi$ has a unique answer set\footnote{Recall that
if $\Pi$ has an inconsistent answer set, we will denote it as
$Lit$. This proposition is a direct generalization of the result for 
general logic programs as described in \cite{gl:stable}.}.
\end{proposition}

Now we define the concept of mutual defeasibility
which plays a key role in investigating a sufficient condition for
the unique answer set of a PLP.

\begin{definition}
Let $\Pi$ be an extended logic program
and $r_{p}$ and $r_{q}$ be two rules in $\Pi$.
We define a set ${\cal D}(r_{p})$ of literals with respect to
$r_{p}$ as follows:
\begin{quote}
${\cal D}_{0}=\{head(r_p)\}$;\\
\hspace*{.1in} ${\cal D}_{i}={\cal D}_{i-1}\cup \{head(r)\mid head(r')\in pos(r)$
where $r\in \Pi$ and 
$r'$ are those\\
\hspace*{1.4in}  rules such that $head(r')\in {\cal D}_{i-1}\}$;\\ 
\hspace*{.1in} ${\cal D}(r_{p})=\bigcup_{i=1}^{\infty} {\cal D}_{i}$.
\end{quote}
We say that
$r_{q}$ is {\em defeasible through} $r_{p}$ in $\Pi$ if and only if
$neg(r_{q})\cap {\cal D}(r_{p})\neq \emptyset$.
$r_{p}$ and $r_{q}$ are called {\em mutually defeasible} in $\Pi$ if
$r_{q}$ is defeasible through $r_{p}$ and
$r_{p}$ is defeasible through $r_{q}$ in $\Pi$.
\end{definition}

Intuitively, if $r_{q}$ is defeasible through $r_{p}$ in $\Pi$, then
there exists a sequence of rules
$r_{1}, r_{2}, \cdots, r_{l}, \cdots$ such that
$head(r_{p})$ occurs in $pos(r_{1})$, $head(r_{i})$ occurs
in $pos(r_{i+1})$ for all $i=1, \cdots$, and for some $k$, $head(r_{k})$ occurs in
$neg(r_{q})$. Under this condition, it is clear that by triggering rule
$r_{p}$ in $\Pi$, it is possible to defeat rule $r_{q}$ if
rules $r_{1}, \cdots, r_{k}$ are triggered as well.
As a special case that ${\cal D}(r_{p})=\{head(r_p)\}$, $r_{q}$ is defeasible
through $r_{p}$ iff $head(r_{p})\in neg(r_{q})$.
The following proposition simply describes the relationship between
local stratification and mutual defeasibility.

\begin{proposition}
Let $\Pi$ be an extended logic program.
If $\Pi$ is locally stratified, then there does not exist
mutually defeasible pair of rules in $\Pi$.
\end{proposition}

The above result is easy to prove from the corresponding result for general
logic programs showed in \cite{gl:stable} based on
Gelfond and Lifschitz's translation
from an extended logic program to a general logic program \cite{gl:logic}.
It is observed that for a PLP ${\cal P}=(\Pi,{\cal N},<)$, if
$\Pi$ is locally stratified, then ${\cal P}$ will have a unique answer set.
In other words, $\Pi$'s local stratification implies that ${\cal P}$
has a unique answer set.
However, this condition seems too strong because
many prioritized logic programs will still have unique answer sets although
their corresponding extended logic programs
are not locally stratified. For
instance, program ${\cal P}_{1}$ presented in section 2 has a unique
answer set but its corresponding
extended logic program is not locally stratified.
But one fact is clear: the uniqueness of reduct for a PLP is necessary
to guarantee this PLP to have a unique answer set.

The above observation suggests that we should first investigate
the condition under which a prioritized logic program has a unique
reduct. Then by applying Proposition 3 to the unique reduct of the PLP,
we obtain the unique answer set condition for this PLP.

\begin{definition}
Let ${\cal P}=(\Pi,{\cal N},<)$ be a PLP.
A $<$-{\em partition} of $\Pi$ in
${\cal P}$ is a finite collection $\{\Pi_{1},\cdots, \Pi_{k}\}$,
where $\Pi=\Pi_{1}\cup\cdots \cup\Pi_{k}$ and
$\Pi_{i}$ and $\Pi_{j}$ are disjoint for any $i\neq j$, such that
\begin{enumerate}
\item
${\cal N}(r)<{\cal N}(r')$ $\in {\cal P}(<)$ implies that
there exist some $i$ and $j$ ($1\leq i<j$) such that
$r'\in \Pi_{j}$ and $r\in \Pi_{i}$;
\item
for each rule $r' \in \Pi_{j}$ ($j>1$),
there exists some rule $r\in \Pi_{i}$ ($1\leq i < j$) such that
${\cal N}(r)<{\cal N}(r') \in {\cal P}(<)$.
\end{enumerate}
\end{definition}

\begin{example}
Consider a PLP ${\cal P}_{4}=(\Pi,{\cal N},<)$:
\begin{quote}
$N_{1}: A\leftarrow$ {\em not} $B$, {\em not} $C$,\\
\hspace*{.1in} $N_{2}: B\leftarrow$ {\em not} $\neg C$,\\
\hspace*{.1in} $N_{3}: C\leftarrow$ {\em not} $A$, {\em not} $\neg C$,\\
\hspace*{.1in} $N_{4}: \neg C\leftarrow$ {\em not} $C$, \\
\hspace*{.1in} $N_{1}<N_{2}, N_{2}<N_{4}, N_{3}<N_{4}$.
\end{quote}
It is easy to verify that a $<$-partition of $\Pi$ in
${\cal P}_{4}$ is $\{\Pi_{1}, \Pi_{2}, \Pi_{3}\}$, where
\begin{quote}
$\Pi_{1}$:  \\
\hspace*{.2in} $N_{1}: A\leftarrow$ {\em not} $B$, {\em not} $C$, \\
\hspace*{.2in} $N_{3}: C\leftarrow$ {\em not} $A$, {\em not} $\neg C$,\\
\hspace*{.1in} $\Pi_{2}$:\\
\hspace*{.2in} $N_{2}: B\leftarrow$ {\em not} $\neg C$,\\
\hspace*{.1in} $\Pi_{3}$:  \\
\hspace*{.2in} $N_{4}: \neg C\leftarrow$ {\em not} $C$.
\end{quote}
In fact, this program has a unique answer set $\{B, C\}$.
\end{example}

\begin{theorem}
Every prioritized logic program has a $<$-partition.
\end{theorem}

\noindent
\begin{proof}
For a given PLP ${\cal P}=(\Pi,{\cal N},<)$,
we construct a series of subsets of $\Pi$ as follows:\\
\hspace*{.1in} $\Pi_{1}=\{r\mid \mbox{there does not exist a rule } r'\in \Pi$
$\mbox{ such that } {\cal N}(r')<{\cal N}(r)\}$;\\
\hspace*{.1in} $\Pi_{i}=\{r\mid \mbox{for all rules such that }$
${\cal N}(r')<{\cal N}(r)$, $r'\in \bigcup_{j=1}^{i-1}\Pi_{j}\}$.\\
We prove that $\{\Pi_{1},\Pi_{2},\cdots\}$ is a
$<$-partition of ${\cal P}$.
First, it is easy to see that $\Pi_{i}$ and $\Pi_{j}$ are disjoint.
Now we show that this partition satisfies conditions 1 and 2 described in
Definition 6. Let ${\cal N}(r)<{\cal N}(r')\in {\cal P}(<)$. If
there does not exist any rule $r''\in \Pi$ such that
${\cal N}(r'')<{\cal N}(r)$, then $r\in\Pi_{1}$. Otherwise, there exists some
$i$ ($i>1$) such that $r\in \Pi_{i}$ and for all rules satisfying
${\cal N}(r'')<{\cal N}(r)$ $r''\in \Pi_{1}\cup\cdots\cup\Pi_{i-1}$.
Let $r'\in \Pi_{j}$. Since ${\cal N}(r)<{\cal N}(r')$, it follows that $1<j$.
From the construction of $\Pi_{j}$, we also conclude
$r\in \Pi_{1}\cup\cdots\cup\Pi_{j-1}$. Since $r'\in \Pi_{j}$,
it follows $i\leq j-1$. That is, $i<j$.
Condition 2 directly follows
from the construction of the partition described above.

Now we show that $\{\Pi_{1},\Pi_{2}, \cdots\}$ must be a finite set.
First, if $\Pi$ is finite, it is clear
$\{\Pi_{1},\Pi_{2}, \cdots\}$ must be a finite set.
If $\Pi$ contains infinite rules, then according to our assumption presented in
section 2,
${\cal P}$ must be the ground instantiation of some program, say
${\cal P}^{*}=(\Pi^{*}, {\cal N}^{*}, <^{*})$.
Then we can use the same way to define
a $<$-partition for ${\cal P}^{*}$. Since
$\Pi^{*}$ is finite, the partition of ${\cal P}^{*}$ must be
also finite: $\{\Pi_{1}^{*}, \Pi_{2}^{*}, \cdots, \Pi_{k}^{*}\}$.
As ${\cal P}^{*}$ is well formed, it implies that for each $i$,
$\Pi_{i}$ is the ground instantiation of $\Pi_{i}^{*}$. So
$\{\Pi_{1},\Pi_{2},\cdots\}$ is finite.
\end{proof}

\begin{theorem}

({\bf Unique Answer Set Theorem})
Let ${\cal P}=(\Pi,{\cal N}<)$ be a PLP and $\{\Pi_{1},\cdots, \Pi_{k}\}$
be a $<$-partition of $\Pi$ in ${\cal P}$.
${\cal P}$ has a unique reduct if there does not exist
two rules $r_{p}$ and $r_{q}$
in $\Pi_{i}$ and $\Pi_{j}$ ($i,j>1$) respectively such that
$r_{p}$ and $r_{q}$ are mutually defeasible in $\Pi$.
${\cal P}$ has a unique answer set if ${\cal P}$ has a unique locally
stratified reduct.
\end{theorem}

\noindent
\begin{proof}
According to Proposition 3, it is sufficient to only prove the
first part of this theorem: ${\cal P}$ has a unique
reduct if there does not exist
two rules $r_{p}$ and $r_{q}$
in $\Pi_{i}$ and $\Pi_{j}$ ($i,j>1$) respectively such that
$r_{p}$ and $r_{q}$ are mutually defeasible in $\Pi$.

We assume that ${\cal P}$ has two different reducts, say
${\cal P}^{<(1)}$ and ${\cal P}^{<(2)}$. This follows that there exist at least
two different rules $r_{p}$ and $r_{q}$ such that (1) $r_{p} \in \Pi_{i}$ and
$r_{q}\in \Pi_{j}$, where $i, j>1$;
(2) $r_{q}\in {\cal P}^{<(1)}$, $r_{q}\not\in {\cal P}^{<(2)}$, and
$r_{p}\not\in {\cal P}^{<(1)}$; and
(3) $r_{p}\in {\cal P}^{<(2)}$, $r_{p}\not\in {\cal P}^{<(1)}$, and
$r_{q}\not\in {\cal P}^{<(2)}$.
According to Definition 2, ${\cal P}^{<(1)}$ and
${\cal P}^{<(2)}$ are generated from two
reduct chains $\{\Pi_{0}^{(1)}, \Pi_{1}^{(1)}, \cdots\}$ and
$\{\Pi_{0}^{(2)}, \Pi_{1}^{(2)}, \cdots\}$ respectively.

Without loss of generality, we may assume that 
for all
$0\leq i < k$, $\Pi_{i}^{(1)}=\Pi_{i}^{(2)}$, and
\begin{quote}
$\Pi_{k}^{(1)}=\Pi_{k-1}^{(1)} - \{r_{1},\cdots, r_{l}, r_{p},\cdots\}$,\\
\hspace*{.1in} $\Pi_{k}^{(2)}=\Pi_{k-1}^{(2)} - \{r_{1},\cdots, r_{l}, r_{q},\cdots\}$,
\end{quote}
where we set $\Pi_{k-1}=\Pi_{k-1}^{(1)}=\Pi_{k-1}^{(2)}$ and the only difference
between $\Pi_{k}^{(1)}$ and $\Pi_{k}^{(2)}$ is due to rules $r_{p}$ and $r_{q}$.
Let $r_{p}$ and $r_{q}$ have the following forms:
\begin{quote}
$r_{p}: L_{p}\leftarrow\cdots$, {\em not} $L_{p}', \cdots$,\\
\hspace*{.1in} $r_{q}: L_{q}\leftarrow\cdots$, {\em not} $L_{q}', \cdots$.
\end{quote}
Comparing $\Pi_{k}^{(1)}$ and $\Pi_{k}^{(2)}$, it is clear that the only difference
between these two programs is about rules $r_{p}$ and $r_{q}$.  Since
$\Pi_{k}^{(1)}$ defeats $r_{p}$ and $\Pi_{k}^{(2)}$ defeats $r_{q}$, it follows that
$L_{q}'\in S_{k}^{(1)}$ and $L_{p}'\in S_{k}^{(2)}$, where
$S_{k}^{(1)}$ and $S_{k}^{(2)}$ are answer sets of
$\Pi_{k}^{(1)}$ and $\Pi_{k}^{(2)}$ respectively.
Then there must exist some rule in $\Pi_{k}^{(1)}$ of the form:
\begin{quote}
$r^{(1)}: L_{p}'\leftarrow\cdots$,
\end{quote}
and some rule in $\Pi_{k}^{(2)}$ of the form:
\begin{quote}
$r^{(2)}: L_{q}'\leftarrow\cdots$.
\end{quote}
Furthermore, since $\Pi_{k}^{(1)}-\{r_{p}, r_{q}\}$
does not defeat rule $r_{p}$ and $\Pi_{k}^{(2)}-\{r_{p}, r_{q}\}$
does not defeat rule $r_{q}$ (otherwise
$\Pi_{k}^{(1)}=\Pi_{k}^{(2)}$), it is observed that
rule $r_{q}$ triggers rule $r^{(1)}$ in $\Pi_{k}^{(1)}$
that defeats $r_{p}$, and rule $r_{p}$
triggers rule $r^{(2)}$ in $\Pi_{k}^{(2)}$
that defeats $r_{q}$. This follows that $r_{p}$ and
$r_{q}$ are mutually defeasible in $\Pi$.
\end{proof}

Note that according to Proposition 4, the condition 
for ${\cal P}=(\Pi,{\cal N},<)$ to have a unique answer set 
stated in Theorem 4 is weaker than the  
local stratification requirement for $\Pi$ to have
a unique answer set as showed by Proposition 3.

\begin{example}
Consider PLP ${\cal P}_{5}=(\Pi, {\cal N}, <)$ as follows:
\begin{quote}
$N_{1}: A\leftarrow$ {\em not} $B$, {\em not} $C$, {\em not} $D$,\\
\hspace*{.1in} $N_{2}: B\leftarrow$ {\em not} $A$, {\em not} $D$, \\
\hspace*{.1in} $N_{3}: C\leftarrow$ {\em not} $A$, {\em not} $D$,\\
\hspace*{.1in} $N_{4}: D\leftarrow$ {\em not} $A$,\\
\hspace*{.1in} $N_{1}<N_{2}<N_{3}$.
\end{quote}
Clearly, a $<$-partition of $\Pi$ is as follows:
\begin{quote}
$\Pi_{1}$:\\
\hspace*{.2in} $N_{1}: A\leftarrow$ {\em not} $B$, {\em not} $C$ {\em not} $D$,\\
\hspace*{.2in} $N_{4}: D\leftarrow$ {\em not} $A$,\\
\hspace*{.1in} $\Pi_{2}$:\\
\hspace*{.2in} $N_{2}: B\leftarrow$ {\em not} $A$, \\
\hspace*{.1in} $\Pi_{3}$:\\
\hspace*{.2in} $N_{3}: C\leftarrow$ {\em not} $A$.
\end{quote}
Although $\Pi$ is not locally stratified, from Theorem 4,
${\cal P}_{5}$ should have a unique reduct $\{N_{1}\}$ since 
$N_{2}$ and $N_{3}$ are not mutually defeasible. This also
concludes that ${\cal P}_{5}$ has a unique answer set $\{A\}$.
\end{example}

\section{Splitting Prioritized Logic Programs}

It has been observed that deciding whether a prioritized logic
program has an answer set is NP-complete \cite{yan:00lp}. That means,
in practice it is unlikely to implement a polynomial algorithm to compute
the answer set of
a prioritized logic program. Hence,
finding suitable strategy to simplify such computation is an important issue.
Similarly to the case of extended logic programs \cite{lt:lp94}, we will show that
under proper conditions, 
a PLP ${\cal P}$ can be split into several smaller components
${\cal P}_{1}, \cdots, {\cal P}_{k}$ such that the evaluation of ${\cal P}$'s
answer sets can be based on the evaluation of the answer sets of
${\cal P}_{1}, \cdots, {\cal P}_{k}$.
To describe our idea, we
first consider the case of splitting a PLP into two parts.

\begin{example}
Consider the following PLP ${\cal P}_{6}=(\Pi, {\cal N}, <)$:
\begin{quote}
$N_{1}: A\leftarrow not \neg A, not D$, \\
\hspace*{.1in} $N_{2}: D\leftarrow not \neg D$,\\
\hspace*{.1in} $N_{3}: \neg D\leftarrow not D$,\\
\hspace*{.1in} $N_{4}: B\leftarrow not C$,\\
\hspace*{.1in} $N_{5}: C\leftarrow not B$, \\
\hspace*{.1in} $N_{6}: A\leftarrow C, \neg D$,\\
\hspace*{.1in} $N_{1}<N_{4}, N_{6}<N_{2}$.
\end{quote}
Clearly, this PLP has a unique reduct $\{N_{1}, N_{3}, N_{5}, N_{6}\}$, which gives
a unique answer set $\{A, C, \neg D\}$.

We observe that $\Pi$ actually can be split into two segments
$\Pi_{1}=\{N_{1}, N_{2}, N_{3}\}$ and $\Pi_{2}=\{N_{4}, N_{5}, N_{6}\}$ such that
$head(\Pi_{2})\cap body(\Pi_{1})=\emptyset$. Now we try to reduce
the computation of ${\cal P}_{6}$'s answer sets to the computation of two
smaller PLPs' answer sets. Firstly, we define a PLP
${\cal P}_{6}^{1}=(\Pi_{1}^{*}, {\cal N}, <)$ by setting
$\Pi_{1}^{*}=\Pi_{1}\cup\{N_{0}: First\leftarrow\}$ and
${\cal P}_{6}^{1}(<)=\{N_{0}<N_{2}\}$.
The role of rule $N_{0}$ is to introduce a $<$-relation
$N_{0}<N_{2}$ to replace the original $<$-relation $N_{6}<N_{2}$ in ${\cal P}_{6}$
that is missed from ${\cal P}_{6}^{1}$ by eliminating $N_{6}$ from $\Pi_{1}$.
The unique answer set of ${\cal P}_{6}^{1}$ is $S_{1}=\{First, A, \neg D\}$.
Since $head(\Pi_{2})\cap body(\Pi_{1})=\emptyset$, it is easy to see that
in each of ${\cal P}_{6}$'s answer sets,
any literals derived by using rules in $\Pi_{2}$ will
not trigger or defeat any rules in $\Pi_{1}$. This implies that every literal in
${\cal P}_{6}^{1}$'s answer set (except {\em First}) will also occur in an answer set of
the original ${\cal P}_{6}$. Therefore, we can define another PLP
${\cal P}_{6}^{2}=(\Pi_{2}^{*}, {\cal N}, <)$ by setting
$\Pi_{2}^{*}$
$=\{N_{4}, N_{5}\}\cup\{N_{0}: First\leftarrow, \hspace*{.05in} N_{6}': A\leftarrow C\}$
and $N_{0}<N_{4}$. Here $N_{6}$ in $\Pi_{2}$ is
replaced by $N_{6}'$ under ${\cal P}_{6}^{1}$'s
answer set $S_{1}$ providing $\neg D$ to be true. Then
${\cal P}_{6}^{2}$ also has a unique answer set $S_{2}=\{First, A, C\}$.
Finally,  the unique answer set of
${\cal P}_{6}$ is obtained by $S_{1}\cup S_{2}-\{First\}=\{A, C, \neg D\}$.
\end{example}

From the above example, we see that if in a PLP ${\cal P}=(\Pi, {\cal N}, <)$,
$\Pi$ can be split into two parts $\Pi_{1}$ and $\Pi_{2}$ such that
$head(\Pi_{2})\cap body(\Pi_{1})=\emptyset$, then it is possible to
also split ${\cal P}$ into two smaller PLPs
${\cal P}^{1}$ and ${\cal P}^{2}$ such that
${\cal P}$'s every answer set can be computed from ${\cal P}^{1}$ and ${\cal P}^{2}$'s.
To formalize our result, we first introduce some useful notions.
Given a PLP ${\cal P}=(\Pi,{\cal N},<)$,
we define a map $first^{<}: \Pi \longrightarrow \Pi$, such that
$first^{<}(r)=r'$ if there exists some
$r'$ such that ${\cal N}(r')<{\cal N}(r)\in {\cal P}(<)$ and there does not exist
another $r'' \in \Pi$ satisfying ${\cal N}(r'')<{\cal N}(r')\in {\cal P}(<)$;
otherwise $first^{<}(r)=undefined$.
Intuitively, $first^{<}(r)$ gives the rule which is most preferred than $r$
in ${\cal P}$. As $<$ is a strict partial ordering,
there may be more than one most preferred rules than $r$.

To define a split of a PLP, we first introduce the concept of $e$-reduct of 
an extended logic program. Let $\Pi$ be an extended logic 
program and $X$ be a set of ground literals.
The $e$-{\em reduct} of $\Pi$
with respect to set $X$ is an extended logic program, denoted as $e(\Pi,X)$, 
obtained from $\Pi$ by deleting (1) each rule in $\Pi$ that has 
a formula $not L$ in its body with $L\in X$, and (2) all
formulas of the form $L$ in the bodies of the remaining rules with $L\in X$.
Consider an example that $X=\{C\}$ and $\Pi$ consists of two rules:
\begin{quote}
$A\leftarrow B, not C$,\\
\hspace*{.1in} $B\leftarrow C, not A$.
\end{quote}
Then $e(\Pi,X)=\{B\leftarrow not A\}$. Intuitively, the $e$-reduct
of $\Pi$ with respect to $X$ can be viewed as a simplified 
program of $\Pi$ given the fact that every literal in $X$ is true.
For a
rule $r\in e(\Pi,X)$, we use
$original(r)$ to denote $r$'s original form in $\Pi$.
In the above example, it is easy to see that
$original(B\leftarrow not A)$ is $B\leftarrow C, not A$.
Now a split of a PLP can be formally defined as follows.

\begin{definition}
Let ${\cal P}=(\Pi, {\cal N},<)$. We say that $({\cal P}^{1}, {\cal P}^{2})$ is
a {\em split} of ${\cal P}$, if there exist two disjoint
subsets $\Pi_{1}$ and $\Pi_{2}$ of
$\Pi$, where $\Pi=\Pi_{1}\cup\Pi_{2}$,  such that
\begin{enumerate}
\item $head(\Pi_{2})\cap body(\Pi_{1})=\emptyset$,
\item ${\cal P}^{1}=(\Pi_{1}\cup\{N_{0}: First\leftarrow\}, {\cal N},<)$\footnote{Here
we assume that {\em First} is a ground literal not occurring in ${\cal P}$.}, where
for any $r, r'\in \Pi_{1}$, ${\cal N}(r)<{\cal N}(r')\in {\cal P}(<)$ implies
${\cal N}(r)<{\cal N}(r')\in {\cal P}^{1}(<)$, and if there exists some
$r''\in \Pi_{2}$ and $first^{<}(r)=r''$, then $N_{0}<{\cal N}(r)\in {\cal P}^{1}(<)$;
\item ${\cal P}^{2}=(e(\Pi_{2},S_{1})\cup\{N_{0}: First\leftarrow\}, {\cal N},<)$, where
$S_{1}$ is an answer set of ${\cal P}^{1}$,
for any $r, r'\in e(\Pi_{2},S_{1})$,
${\cal N}(original(r))<{\cal N}(original(r'))\in {\cal P}(<)$ implies
${\cal N}(r)<{\cal N}(r')\in {\cal P}^{2}(<)$, and if there exists some
$r''\in \Pi_{1}$ and $first^{<}(original(r))$ $=r''$,
then $N_{0}<{\cal N}(r)\in {\cal P}^{2}(<)$.
\end{enumerate}
A split $({\cal P}^{1}, {\cal P}^{2})$ is called $S$-{\em dependent} if
$S$ is an answer set of ${\cal P}^{1}$ and ${\cal P}^{2}$ is formed based on
$S$ as described in condition 3 above,
i.e. ${\cal P}^{2}=(e(\Pi_{2},S)\cup\{N_{0}: First\leftarrow\}, {\cal N},<)$.
\end{definition}

In Example 4, it is easy to verify that
$({\cal P}_{6}^{1},{\cal P}_{6}^{2})$ is a split of ${\cal P}_{6}$.
Now we have the major result of splitting a PLP.

\begin{theorem}
Let $({\cal P}^{1}, {\cal P}^{2})$ be a $S_1$-dependent split of 
${\cal P}$ as defined in Definition 7.  A set of ground literals $S$
is a consistent answer set of ${\cal P}$ if and only if
$S=S_{1}\cup S_{2}-\{First\}$, where 
$S_{2}$ is an answer set of ${\cal P}^{2}$, and
$S_{1}\cup S_{2}$ is consistent.
\end{theorem}

\noindent
\begin{proof}
We prove this theorem in two steps. Suppose $\Pi^{*}$ is a reduct of ${\cal P}$.
According to Definition
2, $\Pi^{*}$ can be represented as the form
$\Pi^{*}=\Pi_{1}^{*}\cup\Pi_{2}^{*}$, where $\Pi_{1}^{*}\subseteq\Pi_{1}$ and
$\Pi_{2}^{*}\subseteq \Pi_{2}$.
So every answer set of $\Pi^{*}$ is also an answer set of ${\cal P}$.
In the first step, 
we prove \underline{\em Result 1}: a set $S$ of ground literals is a
consistent answer set of $\Pi^{*}$ iff $S=S_{1}\cup S_{2}$, where
$S_{1}$ is an answer set of $\Pi_{1}^{*}$, $S_{2}$ is an answer set of
$e(\Pi_{2}^{*},S_{1})$, and $S_{1}\cup S_{2}$ is consistent.
In the second step, we prove \underline{\em Result 2}:
$\Pi_{1}^{*}\cup\{First\leftarrow\}$ is a reduct of
${\cal P}^{1}$ and $e(\Pi_{2}^{*},S_{1})\cup\{First\leftarrow\}$ is a reduct of
${\cal P}^{2}$. Then the theorem is proved  directly from these two results.

We first prove \underline{\em Result 1}. ($\Leftarrow$)
Let $S=S_{1}\cup S_2$ and
$\Pi^{*}=\Pi_{1}^{*}\cup\Pi_{2}^{*}$, where $S_{1}$ is an answer set
of $\Pi_{1}^{*}$ and $S_2$ is an answer set of
$e(\Pi_{2}^{*},S_{1})$ and  $S_{1}\cup S_2$
is consistent. 
Consider the Gelfond-Lifschitz transformation of
$\Pi^{*}$ in terms of $S$, $\Pi^{* S}$.
$\Pi^{* S}$ is obtained from $\Pi^{*}$ by deleting
\begin{enumerate}
\item[(i)] each rule in
$\Pi_{1}^{*}\cup\Pi_{2}^{*}$ that has a formula {\em not} $L$ in its
body with $L\in S$; and
\item[(ii)] all formulas of the form
{\em not} $L$ in the bodies of the remaining rules.
\end{enumerate}
Since $body(\Pi_{1}^{*})\cap head(\Pi_{2}^{*})=\emptyset$, during the
step (i) in the above transformation, for each literal
$L\in S_{1}$, only rules of the
form $L'\leftarrow\cdots$, {\em not} $L, \cdots$
in $\Pi_{1}^{*}$ or $\Pi_{2}^{*}$ will be deleted.
On the other hand, for each literal $L\in S_2$,
only rules of the form $L'\leftarrow\cdots$, {\em not} $L, \cdots$ in
$\Pi_{2}^{*}$ will be deleted and no rules in $\Pi_{1}^{*}$ can be deleted
because $head(\Pi_2^{*})\cap body(\Pi_1^{*})=\emptyset$.
Therefore, we can denote
$\Pi^{* S}$ as $\Pi_{1}^{*'}\cup \Pi_{2}^{*'}$, where
$\Pi_{1}^{*'}$ is obtained from $\Pi_{1}^{*}$ in terms of literals in $S_{1}$, and
$\Pi_{2}^{*'}$ is obtained from $\Pi_{2}^{*}$ in terms of literals in
$S_{1}\cup S_2$
during the above transformation procedure. Then it is easy to see
that $\Pi_{1}^{*'}=\Pi_{1}^{* S_{1}}$. So $S_{1}$ is an answer set of
$\Pi_{1}^{*'}$.

On the other hand, from the construction of $e(\Pi_{2}^{*},S_{1})$,
it is observed that there exists the following correspondence
between $\Pi_{2}^{*'}$ and $e(\Pi_{2}^{*},S_{1})$:
for each rule
\begin{quote}
$L_{0}\leftarrow L_{1},\cdots, L_{k},L_{k+1},\cdots, L_{m}$
\end{quote}
in $\Pi_{2}^{*'}$, there is a rule of the form
\begin{quote}
$L_{0}\leftarrow L_{1},\cdots, L_{k},not L_{m+1},\cdots, not L_{n}$ 
\end{quote}
in $e(\Pi_{2},S_{1})$ such that
$L_{k+1},\cdots, L_{m}\in S_{1}$ and
$L_{m+1},\cdots$, $L_{n}\not\in S_{1}$; on the other hand,
for each rule
$L_{0}\leftarrow L_{1},\cdots, L_{k}$, {\em not} $L_{m+1},\cdots$,
{\em not} $L_{n}$ in $e(\Pi_{2}^{*},S_{1})$,
if none of $L_{m+1},\cdots, L_{n}$ is in $S_2$, then
there exists a rule $L_{0}\leftarrow L_{1},\cdots, L_{k},L_{k+1},\cdots, L_{m}$ in
$\Pi_{2}^{*'}$ such that $L_{k+1},\cdots, L_{m}\in S_{1}$.
From this observation, it can be seen that there
exists a subset $\Delta$ of $S_{1}$ such that
$\Delta\cup S_2$ is an answer set of $\Pi_{2}^{*'}$.
This follows that $S_{1}\cup S_2$ is the smallest set
such that for each rule $L_{0}\leftarrow L_{1},\cdots, L_{m}\in \Pi^{* S}$,
$L_{1},\cdots,L_{m}\in S$ implies $L_{0}\in S$.
That is, $S$ is an answer set of $\Pi^{* S}$ and also
an answer set of $\Pi^{*}$.

($\Rightarrow$) Let $\Pi^{*}=\Pi_{1}^{*}\cup \Pi_{2}^{*}$ and
$S$ be a consistent answer set of $\Pi^{*}$.
It is clear that for each literal $L\in S$, there
must exist some rule of the form $L\leftarrow \cdots$ in $\Pi$.
So we can write $S$ as a form of $S'_{1}\cup S'_{2}$
such that $S'_{1}\subseteq head(\Pi_{1}^{*})$ and
$S'_{2}\subseteq head(\Pi_{2}^{*})$. Note that
$S'_{1}\cap S'_{2}$ may not be empty.
Now we transfer set $S'_{1}$ into $S_{1}$ by the following step:
if $S'_{1}\cap S'_{2}=\emptyset$, then
$S_{1}=S'_{1}$; otherwise, let
\begin{quote}
$S_{1} =S'_{1} -$
$\{L\mid L\in S'_{1}\cap S'_{2}$, and for each rule\\
\hspace*{.9in}
$L\leftarrow L_{1},\cdots,L_{m}$, {\em not}$L_{m+1},\cdots$, {\em not}$L_{n}$ in
$\Pi_{1}^{*}$, there exists some\\
\hspace*{.9in}
$L_{j}$ ($1\leq j\leq m$) $\not\in S'_{1}$ or
$L_{j}\in S'_{1} (m+1\leq j\leq n)\}$.
\end{quote}
In above translation,
since every $L$ deleted from $S_{1}$ is also in $S'_{2}$,
the answer set $S$ of $\Pi^{*}$ can then be expressed as
$S=S_{1}\cup S'_{2}$.
An important fact is observed from the construction of
$S_{1}$:\\
{\bf Fact 1}. $L\in S_{1}$ iff there exists some rule
in $\Pi_{1}^{*}$ of the form
\begin{quote}
$L\leftarrow L_{1},\cdots,L_{m},not L_{m+1},\cdots, not L_{n}$, 
\end{quote}
such that $L_{1},\cdots,L_{m}$
$\in S_{1}$ and $L_{m+1},\cdots$, or $L_{n}$
$\not\in S_{1}$.

Now we prove $S_{1}$ is an answer set of $\Pi_{1}^{*}$.
We do Gelfond-Lifschitz transformation on $\Pi^{*}$ in terms of
set $S=S_{1}\cup S'_{2}$. After such transformation,
we can write $\Pi^{* S}$ as form
$\Pi_{1}^{*'}\cup\Pi_{2}^{*'}$, where
$\Pi_{1}^{*'}\subseteq \Pi_{1}^{*}$ and $\Pi_{2}^{*'}\subseteq \Pi_{2}^{*}$.
As $head(\Pi_{2}^{*}) \cap body(\Pi_{1}^{*})=\emptyset$,
any literal
in $S'_{2}$ will not cause a deletion of a rule from $\Pi_{1}^{*}$ in
the Gelfond-Lifschitz transformation.
Then it is easy to see that $\Pi_{1}^{*'} =\Pi_{1}^{* S_{1}}$.
From {\bf Fact 1}, it concludes that
literal $L\in S_{1}$ iff there is a rule
$L\leftarrow L_{1},\cdots L_{m}$ in $\Pi_{1}^{* S_{1}}$ and
$L_{1},\cdots,L_{m} \in S_{1}$. This follows that
$S_{1}$ is an answer set of $\Pi_{1}^{* S_{1}}$, and then
an answer set of $\Pi_{1}^{*}$.

Now we transfer $S'_{2}$ into $S_2$ by the following step:
if $S_{1}\cap S'_{2}=\emptyset$, then
$S_2=S'_{2}$; otherwise, let
\begin{quote}
$S_2=S'_{2}-\{L\mid L\in S_{1}\cap S'_{2}$,
and for each rule \\
\hspace*{.9in} $L\leftarrow L_{1},\cdots,L_{m}$,
{\em not} $L_{m+1},\cdots$, {\em not} $L_{n}$ in $\Pi_{2}^{*}$, there exists
some \\
\hspace*{.9in} $L_{j} (1\leq j\leq m) \not\in S_{1}\cup S'_{2}$, or
$L_{j}\in S_{1}\cup S'_{2} (m+1\leq j\leq n)\}$.
\end{quote}
After this translation, $S$ can be expressed as
$S=S_{1}\cup S_2$.
An important fact is also observed from the translation of $S_{2}$:\\
{\bf Fact 2}. $L\in S_2$ iff
there exists some rule in $\Pi_{2}^{*'}$ of the form
\begin{quote}
$L\leftarrow L_{1},\cdots,L_{k},L_{k+1},\cdots,L_{m}$
\end{quote}
such that $L_{1},\cdots,L_{k}\in S_{1}$ and
$L_{k+1},\cdots,L_{m}\in S_2$.

Now we prove $S_2$ is an answer set of $e(\Pi_{1}^{*}, S_{1})$.
Recall that
$\Pi^{* S}=\Pi_{1}^{*'}\cup \Pi_{2}^{*'}=\Pi_{1}^{* S_{1}}\cup \Pi_{2}^{*'}$.
From {\bf Fact 2}, it is clear that there exists a subset $\Delta$ of
$S_{1}$ such that $S_2$ is an answer set of
$e(\Pi_{2}^{*'},\Delta)$ and $e(\Pi_{2}^{*'},\Delta)^{S_2}=e(\Pi_{2}^{*'},\Delta)$.
On the other hand, from the construction
of $e(\Pi_{2}^{*},S_{1})$, it is easy to see that
$e(\Pi_{2}^{*},S_{1})^{S_2}=e(\Pi_{2}^{*'},\Delta)$
$=e(\Pi_{2}^{*'},\Delta)^{S_2}$. So $S_2$ is also an answer set
of $e(\Pi_{2}^{*},S_{1})$.

\comment{

(Sketch)
Under the condition of $head(\Pi_{2}^{*})$
$\cap$ $body(\Pi_{1}^{*})=\emptyset$,
we apply
Gelfond-Lifschitz transformation on $\Pi^{*}$ in terms of $S$, denoted as
$\Pi^{* S}$.
Then we observe that for each $L\in S_{1}$, only rules
with the form $L'\leftarrow\cdots, not L$ in
$\Pi_{1}^{*}$ will be deleted, and for each $L\in S_{2}$, only rules
with the form $L'\leftarrow\cdots, not L$ in $\Pi_{2}^{*}$ will be deleted,
and no other rules in
$\Pi_{1}^{*}$ can be deleted. So we can represent
$\Pi^{* S}$ as $\Pi_{1}^{*'}\cup \Pi_{2}^{*'}$, where $\Pi_{1}^{*'}$ is obtained
from $\Pi_{1}^{*}$ in terms of literals in $S_{1}$, and $\Pi_{2}^{*'}$ is obtained from
$\Pi_{2}^{*}$ in terms of literals in $S_{1}\cup S_{2}$ during the transformation.
Then it is easy to see that $\Pi_{1}^{*'}=\Pi_{1}^{*S_{1}}$. So $S_{1}$ is also an answer
set of $\Pi_{1}^{*'}$. From this fact, we can show that
$S_{1}\cup S_{2}$ is the smallest set such that for each
$L_{0}\leftarrow L_{1},\cdots,L_{m}\in \Pi^{* S}$,
$L_{1},\cdots, L_{m}\in S$ implies $L_{0}\in S$. That is, $S$ is an answer set of
$\Pi^{* S}$, and also an answer set of $\Pi^{*}$.

($\Rightarrow$) Suppose $S$ is a consistent answer set of $\Pi^{*}$.
Let $S=S_{1}'\cup S_{2}'$ where $S_{1}'\subseteq head(\Pi_{1}^{*})$ and
$S_{2}'\subseteq head(\Pi_{2}^{*})$. Note that $S_{1}'\cap S_{2}'$ may not be empty. Now
we transfer $S_{1}'$ to $S_{1}$ as follows: if
$S_{1}'\cap S_{2}'=\emptyset$, then $S_{1}'=S_{1}$; otherwise
$S_{1}=S_{1}'-\{L\mid S_{1}'\cap S_{2}'$, and for each
rule
$L\leftarrow L_{1},\cdots, L_{m}, not L_{m+1},\cdots, not L_{n} \in \Pi_{1}^{*}$, there
exists some $L_{j} (1\leq j\leq m) \not\in S_{1}'$ or
$L_{j} (m+1\leq j\leq n) \in S_{1}'\}$.  Note that in this translation, every
literal deleted from $S_{1}'$ is also in $S_{2}'$.
So $S$ can be also represented as the form $S=S_{1}\cup S_{2}'$. It can be verified that
$S_{1}$ is indeed an  answer set of $\Pi_{1}^{*}$. Similarly, we transfer $S_{2}'$ to
$S_{2}$ as follows: if $S_{1}\cap S_{2}'=\emptyset$, then
$S_{2}=S_{2}'$; otherwise, let
$S_{2}=S_{2}'-\{L\mid L\in S_{1}\cap S_{2}'$, and for each rule
$L\leftarrow L_{1},\cdots, L_{m}, not L_{m+1},\cdots, not L_{n}\in \Pi_{2}^{*}$,
there exists some $L_{j} (1\leq j\leq m)\not\in S_{1}\cup S_{2}'$, or
$L_{j} (m+1\leq j\leq n)\in S_{1}\cup S_{2}'\}$.
After this translation, $S$ can be represented as
$S=S_{1}\cup S_{2}$. Then it can be also verified
that $S_{2}$ is an answer set of $e(\Pi_{2}^{*},S_1)$.

}

Now we show \underline{\em Result 2}. 
The fact that $\Pi_{1}^{*}\cup \{First\leftarrow\}$ is a reduct of ${\cal P}^{1}$ is proved
based on a construction of a 1-1 correspondence between the computation of
${\cal P}$'s reduct and ${\cal P}^{1}$'s reduct.
Let $\{\Pi^{i}\}$ ($i=0, 1, \cdots$)
be the series generated by computing ${\cal P}$'s reduct
(see Definition 2), and $\{\Pi'^{i}\}$ ($i=0, 1, \cdots$) be the series
generated by computing ${\cal P}^{1}$'s reduct. From the specification of
${\cal P}^{1}$ and condition $head(\Pi_{2})\cap body(\Pi_{1})=\emptyset$,
we observe that for each $\Pi^{i}$, which is obtained from
$\Pi^{i-1}$ by eliminating some rules from $\Pi^{i-1}$,
if some rule in $\Pi_{1}$ is deleted, then this rule must be also deleted in
$\Pi'^{i}$; if no rule in  $\Pi_{1}$ is deleted (e.g. all rules deleted
from $\Pi^{i-1}$ are in $\Pi_{2}$),
then we set $\Pi'^{i}=\Pi'^{i-1}$. Then it is clear that
every rule in $\Pi_{1}^{*}\cup \{First\leftarrow\}$ 
must be also in the reduct of ${\cal P}^{1}$ and
{\em vice versa}. This concludes that $\Pi_{1}^{*}\cup\{First\leftarrow\}$ 
is a reduct of ${\cal P}^{1}$.
Similarly we can show that $e(\Pi_{2}^{*},S_{1})\cup \{First\leftarrow\}$ 
is a reduct of ${\cal P}^{2}$.
\end{proof}

Once a PLP has a split, by applying Theorem 5, 
we eventually reduce the computation of a large PLP's answer sets
to the computation of two smaller PLPs' answer sets.
In a general case, it is also possible to split a large PLP into
a series of smaller PLPs.

\begin{definition}
Let ${\cal P}=(\Pi, {\cal N},<)$. We say that $({\cal P}^{1}, \cdots, {\cal P}^{k})$ is
a {\em split} of ${\cal P}$,
if there exist $k$ disjoint subsets $\Pi_{1},\cdots, \Pi_{k}$ of
$\Pi$, where $\Pi=\bigcup_{i=1}^{k}\Pi_{i}$,  such that
\begin{enumerate}
\item $head(\Pi_{i})\cap body(\bigcup_{j=1}^{i-1}\Pi_{j})=\emptyset$, ($i=2, \cdots, k$),
\item ${\cal P}^{1}=(\Pi_{1}\cup\{N_{0}: First\leftarrow\}, {\cal N},<)$, where
for any $r, r'\in \Pi_{1}$, ${\cal N}(r)<{\cal N}(r')\in {\cal P}(<)$ implies
${\cal N}(r)<{\cal N}(r')\in {\cal P}^{1}(<)$, and if there exists some
$r''\not\in \Pi_{1}$ and $first^{<}(r)=r''$, then $N_{0}<{\cal N}(r)\in {\cal P'}(<)$;
\item ${\cal P}^{i}=(e(\Pi_{i},\bigcup_{j=1}^{i-1} S_{j})$
$\cup\{N_{0}: First\leftarrow\}, {\cal N},<)$, where
$S_{j}$ is an answer set of ${\cal P}^{j}$,
for any $r, r'\in e(\Pi_{i},\bigcup_{j=1}^{i-1} S_{j})$,
${\cal N}(original(r))<{\cal N}(original(r'))\in {\cal P}(<)$ implies
${\cal N}(r)<{\cal N}(r')\in {\cal P}^{i}(<)$, and if there exists some
$r''\not\in \Pi_{i}$ and $first^{<}(original(r))$ $=r''$,
then $N_{0}<{\cal N}(r)\in {\cal P}^{i}(<)$.
\end{enumerate}
A split $({\cal P}^{1},\cdots, {\cal P}^{k})$ is called
$\bigcup_{i=1}^{k-1}S_i$-{\em dependent} if
$S_i$ is an answer set of ${\cal P}^{i}$ ($i=1,\cdots,k-1$) 
and each ${\cal P}^{i+1}$ is formed based on 
$\bigcup_{j=1}^{i}S_j$ as described in condition 3 above.
\end{definition}

Now using a similar technique as described in the proof of Theorem 5, we
have the following general splitting result.

\begin{theorem}
Let $({\cal P}^{1}, \cdots, {\cal P}^{k})$ be a 
$\bigcup_{i=1}^{k-1}S_i$-{\em dependent} split of ${\cal P}$ as defined in
Definition 8.  A set of ground literals $S$
is a consistent answer set of ${\cal P}$ if and only if
$S=\bigcup_{i=1}^{k} S_{i}-\{First\}$, where $S_{k}$ is an answer set of
${\cal P}^{k}$, and $\bigcup_{i=1}^{k} S_{i}$ is consistent.
\end{theorem}

\begin{example}
Consider PLP ${\cal P}_{7}=(\Pi, {\cal N},<)$ as follows:
\begin{quote}
$N_{1}: A\leftarrow not B$,\\
\hspace*{.1in} $N_{2}: B\leftarrow not A$,\\
\hspace*{.1in} $N_{3}: C\leftarrow not \neg C$, \\
\hspace*{.1in} $N_{4}: D\leftarrow not B$,\\
\hspace*{.1in} $N_{5}: \neg D\leftarrow not D$,\\
\hspace*{.1in} $N_{1}<N_{2}<N_{3}<N_{4}<N_{5}$.
\end{quote}
Let $\Pi_{1}=\{N_{1}, N_{2}\}$, $\Pi_{2}=\{N_{3}, N_{4}\}$, and $\Pi_{3}=\{N_{5}\}$.
Clearly, $head(\Pi_{2})\cap body(\Pi_{1})=\emptyset$ and
$head(\Pi_{3})\cap body(\Pi_{1}\cup \Pi_{2})=\emptyset$. Then a split
of ${\cal P}_{7}$, denoted as
$({\cal P}_{7}^{1}, {\cal P}_{7}^{2}, {\cal P}_{7}^{3})$,
can be constructed. Ignoring the detail,
this split is illustrated as follows:
\begin{quote}
${\cal P}_{7}^{1}$: \hspace*{1.in}
${\cal P}_{7}^{2}$: \hspace*{1.in} ${\cal P}_{7}^{3}$:\\
\hspace*{.1in} $N_{0}: First\leftarrow$,  \hspace*{.45in} $N_{0}: First\leftarrow$,
\hspace*{.5in} $N_{0}: First\leftarrow$, \\
\hspace*{.1in} $N_{1}: A\leftarrow not B$,
\hspace*{.3in} $N_{3}: C\leftarrow not \neg C$, \\
\hspace*{.1in} $N_{2}: B\leftarrow not A$,
\hspace*{.3in} $N_{4}: D\leftarrow not B$,\\
\hspace*{.1in} $N_{1}<N_{2}$,
\hspace*{.65in} $N_{0}<N_{3}<N_{4}$,
\end{quote} 
Then according to Theorem 6,
each answer set of ${\cal P}_{7}$
can be represented as $S_{1}\cup S_{2}\cup S_{3} - \{First\}$,
where $S_{1}$, $S_{2}$, and $S_{3}$ are answer sets of
${\cal P}_{7}^{1}$, ${\cal P}_{7}^{2}$ and
${\cal P}_{7}^{3}$ respectively, which are
$\{First, A\}$, $\{First, C, D\}$ and $\{First\}$ respectively.
So $\{A, C, D\}$ is the unique answer set of ${\cal P}_{7}$.
\end{example}

\section{Conclusion}

In this paper, we have proved two major results for 
prioritized logic programs:
the unique answer set theorem and
splitting theorems for prioritized logic programs.
By introducing the concept of mutual defeasibility,
the first result provides a sufficient condition for the unique answer 
set of a prioritized logic program. It is observed that the sufficient condition
in Theorem 4 is weaker than the local stratification as required for
extended logic programs.

Our splitting theorems, on the other hand, 
illustrated that as in the case of extended logic programs, under  
certain
conditions, the computation of answer sets of a prioritized logic 
program can be simplified.
It is interesting to note that by omitting preference relation $<$,
our splitting theorems actually also present new results for splitting usual 
extended logic program which 
generalizes Lifschitz and Turner's result \cite{lt:lp94}. Consider
an extended logic program $\Pi$ consisting of the following rules:
\begin{quote}
$A\leftarrow$ {\em not} $C$,\\     
\hspace*{.1in} $A\leftarrow$ {\em not} $B$,\\
\hspace*{.1in} $B\leftarrow$ {\em not} $A$. 
\end{quote}        
This program does not have a non-trivial split under
Lifschitz and Turner's Splitting Set Theorem,
but under our splitting theorem condition, $\Pi$ can 
be split into $\Pi_{1}$ and $\Pi_{2}$ as follows:
\begin{quote}                      
$\Pi_{1}$: \hspace*{1.2in} $\Pi_{2}$:\\
\hspace*{.1in} $A\leftarrow$ {\em not} $C$,
\hspace*{.7in} $A\leftarrow$ {\em not} $B$,\\
\hspace*{1.55in} $B\leftarrow$ {\em not} $A$,
\end{quote}                        
such that $body(\Pi_{1})\cap head(\Pi_{2})=\emptyset$. It is observed that
$\{A\}$ is the unique answer set of $\Pi_{1}$, and
the unique answer set of $\Pi$ is then obtained from
$\Pi_{1}$'s answer set $\{A\}$ and the answer set of
$e(\Pi_{2},\{A\})$, which is also $\{A\}$. So we get the unique answer set
of $\Pi$ $\{A\}$.   A detailed discussion on the relationship between 
Lifschitz and Turner's Splitting Set Theorem and our splitting result on 
extended  logic programs 
is referred to the author's another paper \cite{yan:iclp99}. 


We should state that
our results proved in this paper 
are based on a specific formulation of prioritized logic programming, and hence
it is not clear yet whether they are generally suitable for other prioritized 
default reasoning systems, e.g. \cite{d:plp,g:97,w:plp}. 
However, since the traditional
answer set semantics was employed in our development of 
prioritized logic programming, we would expect that our results could be extended 
to other answer set semantics based PLP frameworks. This will be
an interesting topic for our future work.

Finally, 
it is also worth mentioning that besides the idea of developing
a ``prioritized version'' of answer set semantics for PLP (like the approach 
we discussed in this paper),
there are other approaches to PLP
in which the semantics of PLP is defined by 
modular and simple translation of PLP programs into standard logic
programs. Work on this direction is due to 
Gelfond and Son \cite{ms:98}, Delgrande, Schaub and Tompits \cite{d:plp} and
some other researchers. Recently, Schaub and Wang
further investigated a series of uniform characterizations among these
approaches \cite{sw:01}. For these approaches, the classical splitting
set theorem \cite{lt:lp94} can be used to simplify the reasoning 
procedure of PLP.

\bibliography{final-tworesultR5}

\comment{

}

\end{document}